\documentclass[letterpaper]{article} % DO NOT CHANGE THIS
\usepackage{aaai2027}  % DO NOT CHANGE THIS
\usepackage[hyphens]{url}  % DO NOT CHANGE THIS
\usepackage{graphicx} % DO NOT CHANGE THIS
\usepackage{natbib}  % DO NOT CHANGE THIS AND DO NOT ADD ANY OPTIONS TO IT
\usepackage{caption} % DO NOT CHANGE THIS AND DO NOT ADD ANY OPTIONS TO IT
\usepackage{algorithm}
\usepackage{algorithmic}
\usepackage{amsmath}
\usepackage{multirow} % tables (some linters require this)
\usepackage{pifont} % for checkmarks
\usepackage{amssymb} % for \checkmark
\usepackage{newfloat}
\usepackage{listings}
\DeclareCaptionStyle{ruled}{labelfont=normalfont,labelsep=colon,strut=off} % DO NOT CHANGE THIS
\floatstyle{ruled}
\newfloat{listing}{tb}{lst}{}
\floatname{listing}{Listing}
\usepackage{booktabs}

\author{
    Kai Li\textsuperscript{\rm 1,\rm 2},
    Lutao Jiang\textsuperscript{\rm 3},
    Zhenyang Li\textsuperscript{\rm 4},
    Jiayu Dong\textsuperscript{\rm 5},
    Jierui Zhang\textsuperscript{\rm 4},
    Yingda Yin\textsuperscript{\rm 5},
    Runze Zhang\textsuperscript{\rm 5},
    Kai Yan\textsuperscript{\rm 5},
    Xiaoyang Huang\textsuperscript{\rm 5},
    Keyang Luo\textsuperscript{\rm 5},
    Xin Wang\textsuperscript{\rm 5},
    Xiangyu Zhao\textsuperscript{\rm 1},
    Weikai Chen\textsuperscript{\rm 6}*\\
}
\affiliations{
    \textsuperscript{\rm 1}City University of Hong Kong, Hong Kong SAR, China\\
    \textsuperscript{\rm 2}University of Chinese Academy of Sciences, Beijing, China\\
    \textsuperscript{\rm 3}The Hong Kong University of Science and Technology (Guangzhou), Guangzhou, China\\
    \textsuperscript{\rm 4}The University of Hong Kong, Shenzhen, China\\
    \textsuperscript{\rm 5}LIGHTSPEED, Shenzhen, China\\
    \textsuperscript{\rm 6}LIGHTSPEED, Los Angeles, USA\\
}

\usepackage{threeparttable}
\graphicspath{{figures/}}

\newcommand{\method}{\textsc{Scenix}}
\newcommand{\dataset}{\textsc{XScene}}

\newcommand{\yes}{\checkmark}
\newcommand{\no}{$\times$}
\newcommand{\opt}{$\ast$}
\graphicspath{{figures/}}

\makeatletter
\g@addto@macro\@maketitle{%
  \par
  \begin{minipage}{\textwidth}
    \centering
    \includegraphics[width=\linewidth]{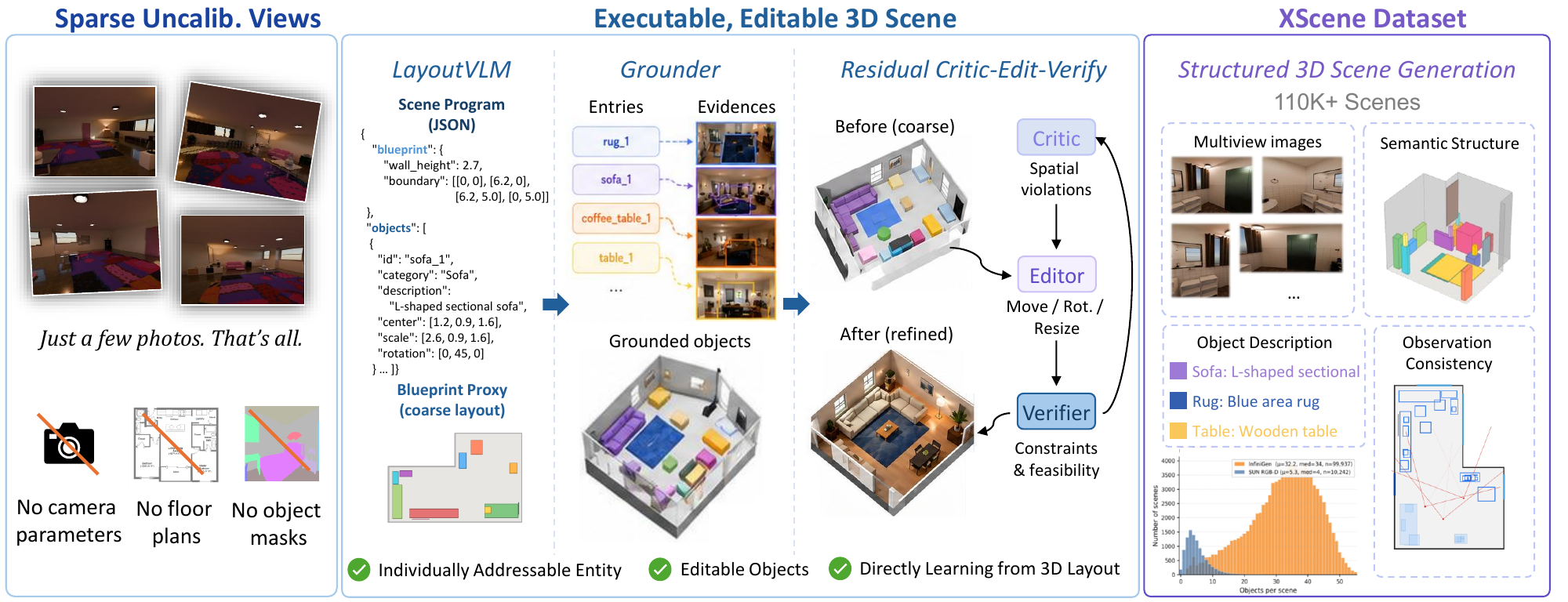}
    \captionof{figure}{%
        \method{} directly transforms sparse uncalibrated RGB observations
        into an executable 3D scene program and realizes it as a structured,
        editable indoor environment.}
    \label{fig:teaser}
\end{minipage}
  \par}
\makeatother

\title{Scenix: Sparse-View 3D Scene Reconstruction via Executable Scene Programs}

\begin{document}

\maketitle
\begin{abstract}
Synthesizing a structured and editable 3D indoor scene from a few uncalibrated RGB views requires more than generating high-quality individual assets: a system must infer the room structure, associate objects across incomplete observations, and recover a globally consistent spatial configuration. 
Previous methods mainly focus on 3D scene generation with text input or require continuous visual inputs with additional priors, \ e.g., human-annotated masks or accurate 3D layouts, which makes these methods labor demanding and hard to apply in general cases. 
We present \method, a sparse-view 3D scene reconstruction framework via executable scene programs, a structured representation that can be directly instantiated into editable 3D scenes. 
Given sparse views, \method{} predicts executable scene programs through perception-grounded asset instantiation and closed-loop spatial refinement.
% We present \method, a framework that predicts an executable scene representation from sparse views and realizes it through perception-grounded asset instantiation and closed-loop spatial refinement. 
To support this task, we construct \dataset, a dataset of approximately 110,000 synthetic and real indoor scenes with multiview imagery, room structures, object-centric descriptions, and metric spatial annotations. We further introduce observation-consistent supervision that aligns each target scene with the visual evidence available in its input views. Experiments on held-out \dataset{} scenes, real indoor images, and out-of-distribution SpatialGen cases evaluate structured scene prediction, object grounding, and spatial refinement.

\end{abstract}

% Keep optional anonymous links between the abstract and main body.
% \begin{links}
%     \link{Code}{https://anonymous.example/code}
%     \link{Dataset}{https://anonymous.example/dataset}
% \end{links}

\section{Introduction}
\label{sec:introduction}

Recent generative models create high-quality 3D assets from a single image~\cite{survey}, yet reconstructing a complete, editable indoor scene from a few uncalibrated RGB views remains harder. Beyond recovering objects, a system must infer room structure, establish correspondences across sparse observations, and recover a globally consistent spatial configuration. 
% These coupled structural and semantic problems distinguish scene reconstruction from object generation.
These tightly coupled structural and semantic reasoning problems make sparse-view scene reconstruction fundamentally more difficult than single-object generation.

Existing paradigms address only part of this problem. Text-conditioned methods~\cite{chen2024spatialvlm,wan2026nala,feng2023layoutgpt,schneider_worldmesh_2026} generate plausible indoor scenes but cannot faithfully reconstruct a particular environment. Neural reconstruction methods recover scene geometry from images, yet typically require dense calibrated views or auxiliary geometric priors and produce representations that are difficult to edit at the object level.
Recent open-vocabulary asset generators~\cite{ye2025hi3dgen,chen2026sam3d,hunyuan3d2025hunyuan3d,xiang2025trellis} have largely removed the restriction of predefined asset libraries. The remaining challenge is no longer asset creation, but \emph{recovering a structured scene description} that guides \emph{how those assets are instantiated and composed}.

% Existing paradigms address only parts of this problem. Text-conditioned systems~\cite{chen2024spatialvlm,wan2026nala,feng2023layoutgpt,schneider_worldmesh_2026} synthesize plausible scenes but cannot reproduce a particular room, while neural reconstruction commonly assumes dense calibrated views or auxiliary geometry and yields representations that are difficult to edit by object. Object-centric pipelines improve controllability through detection, depth or pose estimation, cross-view association, asset retrieval, or other external information~\cite{yan_psdr-room_2023,chen2026sam3d,ardelean_gen3dsr_2025,yin2026threedfixer,ng2024partcraft}, but accumulate errors across stages and remain bounded by predefined libraries. Recent open-vocabulary asset generators~\cite{ye2025hi3dgen,chen2026sam3d,hunyuan3d2025hunyuan3d,xiang2025trellis} relax the library constraint without resolving scene composition; however, each object must still be identified, scaled, oriented, and placed consistently.

% We formulate sparse-view indoor scene synthesis as direct prediction and realization of an executable \emph{structured scene representation}, or 3D scene program. 

This observation suggests that scene understanding should precede object realization. We therefore formulate sparse-view 3D scene reconstruction as executable scene program prediction. 
Rather than reconstructing monolithic geometry, it explicitly encodes the room envelope together with object identities, descriptions, positions, extents, orientations, and support relations. As an interface between perception and realization, it can be learned from images, instantiated with generated assets, verified through canonical projections, and edited at the object level.

\begin{table}[t]
    \centering
    \begingroup
    % \centering
    \scriptsize
    \setlength{\tabcolsep}{2.5pt}
    \renewcommand{\arraystretch}{0.9}
    % tighten booktabs rule spacing
    \setlength{\aboverulesep}{0pt}
    \setlength{\belowrulesep}{0pt}
    \setlength{\cmidrulesep}{0pt}
    \caption{Comparison of input modalities and auxiliary conditions. $\ast$ The VLM framework supports text, image, and video inputs.}
    \resizebox{0.8\columnwidth}{!}{%
    \begin{tabular}{lccccc}
        \toprule
        \textbf{Method} & \textbf{Single} & \textbf{Multi.} & \textbf{Text} & \textbf{Mask} & \textbf{Conditions} \\
        \midrule
        Gen3DSR    & \yes & \no & \no & \yes & Depth{+}Camera \\
        3DFixer    & \yes & \no & \no & \yes & Depth \\
        SAM3D      & \yes & \no & \no & \yes & Diff. Rendering \\
        SpatialGen & \yes & \yes & \no & \no & 3D Layout \\
        NaLA       & \no & \no & \yes & \no & N/A \\
        LayoutGPT  & \no & \no & \yes & \no & N/A \\
        Ours       & \yes & \yes & \opt & \no & N/A \\
        \bottomrule
    \end{tabular}%
    }
    \label{tab:method_comparison}
    \endgroup
    % \vspace{-0.3cm}
\end{table}

\begin{table}[t]
    \centering
    \caption{Comparison of 3D datasets. BP: blueprint; P.\ \&\ E.: positions and extents. $^{\dagger}$3D-FRONT images are rendered from provided scenes.}
    \scriptsize
    \setlength{\tabcolsep}{2.5pt}
    \renewcommand{\arraystretch}{0.95}
    \resizebox{0.8\columnwidth}{!}{%
    \begin{tabular}{l c r r c c c c}
    \toprule
    \textbf{Dataset} & \textbf{Src.} & \textbf{\#Scenes }& \textbf{\#Imgs} & \textbf{BP} &\textbf{ P.\ \&\ E.} & \textbf{Ori. }& \textbf{Desc.}\\
    \midrule
    SUN RGB-D      & real & --     & 10.3K  & \yes & \yes & \no  & \no  \\
    ScanNet        & real & 1,513  & 2.5M   & \no  & \yes & \no  & \no  \\
    Matterport3D   & real & 90     & 10.8K  & \no  & \yes & \no  & \no  \\
    ScanNet++      & real & 1,006  & 11.1M  & \no  & \yes & \no  & \no  \\
    \midrule
    Structured3D   & syn. & 3,500  & 196.5K & \yes & \yes & \no  & \no  \\
    Hypersim       & syn. & 461    & 77.4K  & \no  & \yes & \no  & \no  \\
    SpatialGen     & syn. & 12,328 & 4.7M   & \yes & \yes & \no  & \no  \\
    3D-Front$^{\dagger}$ & syn. & 18,968 & --    & \yes & \yes & \yes & \no  \\
    \midrule
    \dataset{}(Ours)    & mix  & 110K   & 560K    & \yes & \yes & \yes & \yes \\
    \bottomrule
    \label{tab:dataset_compare}
\end{tabular}%
}
\end{table}

Based on this formulation, \method{} reconstructs structured 3D scenes from uncalibrated RGB observations without external camera calibration, point clouds, floor plans, or manually annotated masks. LayoutVLM performs multimodal reasoning over a variable number of views to predict a scene program encoding room layout, object-level spatial attributes, and rich semantic descriptions. Asset Grounder realizes its entities with open-vocabulary 3D assets, followed by bounded Critic--Editor--Verify refinement that preserves structural consistency.

% Based on this formulation, we introduce \method{}, a framework for structured 3D indoor scene synthesis from sparse uncalibrated RGB views. \method{} requires no externally provided camera calibration, depth, point clouds, floor plans, or manually annotated masks. In Tab.~\ref{tab:method_comparison}, we summarizes representative input and supervision requirements comparing across methods. Its first component, \emph{LayoutVLM}, maps a variable number of RGB observations to an executable scene program containing the room blueprint, object-level spatial attributes, as well as rich textual descriptions. Instead of aggregating observations into an implicit geometric field, LayoutVLM directly performs multimodal, scene-level reasoning over a compositional output space.

Training executable scene programs requires supervision unavailable in existing indoor datasets. We therefore construct the \dataset{} dataset, comprising approximately 110,000 synthetic and real indoor scenes with room layouts, object descriptions, and metric spatial annotations. The synthetic portion is procedurally generated with InfiniGen~\cite{raistrick_infinite_2023} to provide scalable multiview supervision, while approximately 10,000 SUN RGB-D~\cite{song2015sunrgbd} images improve real-world coverage. To faithfully model sparse-view reconstruction, we further introduce an observation-consistent annotation protocol that retains only entities visible in the selected observations. Tab.~\ref{tab:dataset_compare} compares \dataset{} with existing indoor datasets~\cite{song2015sunrgbd,dai2017scannet,chang2017matterport3d,yeshwanth2023scannetpp,zheng2020structured3d,roberts2021hypersim,fang2026spatialgen,fu2021threedfront}.

% The predicted program is further realized through two supporting stages. Asset Grounder reconciles observed instances with generated entities and derives object-centric visual conditions for open-vocabulary 3D generation. A residual Critic--Editor--Verify loop then proposes and validates bounded spatial corrections under deterministic geometric constraints. 

Extensive experiments on held-out \dataset{} scenes, comparisons with single-view reconstruction baselines, and zero-shot transfer to SpatialGen demonstrate the effectiveness of our formulation. \method{} achieves more accurate scene layout reconstruction than existing approaches and consistently benefits from grounded residual refinement under distribution shift. Our contributions are threefold:
\begin{itemize}
    \item We introduce \method{}, a new paradigm for sparse-view 3D scene reconstruction that directly infers executable structured scene programs from uncalibrated RGB images.
    \item We construct \dataset, approximately 110,000 indoor cases with appearance-rich descriptions, complete placement parameters, and observation-consistent supervision.
    \item We demonstrate that executable scene programs serve as an effective representation for sparse-view reconstruction, yielding more accurate scene layouts, stronger generalization, and object-level editable 3D scenes.
\end{itemize}

% We evaluate \method{} on held-out \dataset{} scenes, real indoor images, and SpatialGen under out-of-distribution settings. Our experiments assess structured scene prediction, object grounding, spatial consistency, and the effect of closed-loop refinement.

% Our contributions are threefold:
% \begin{itemize}
%     \item We introduce \method, which formulates sparse-view indoor scene synthesis as the prediction, grounding, and realization of an explicit structured scene representation, without requiring externally provided geometric information.
%     \item We construct \dataset, a dataset of approximately 110,000 indoor scenes with multiview imagery, room structures, object-centric descriptions, and metric spatial annotations, together with observation-consistent supervision aligned with visible evidence.
%     \item We propose perception-grounded asset instantiation and closed-loop spatial refinement, bridging structured prediction and open-world 3D realization while preserving object-level editability.
% \end{itemize}

% First ordinary figure* in the paper: LaTeX places it at the top of page 2.
% Replace only the fbox with the final \includegraphics command.
\begin{figure*}[t]
    \centering
    \includegraphics[width=\linewidth]{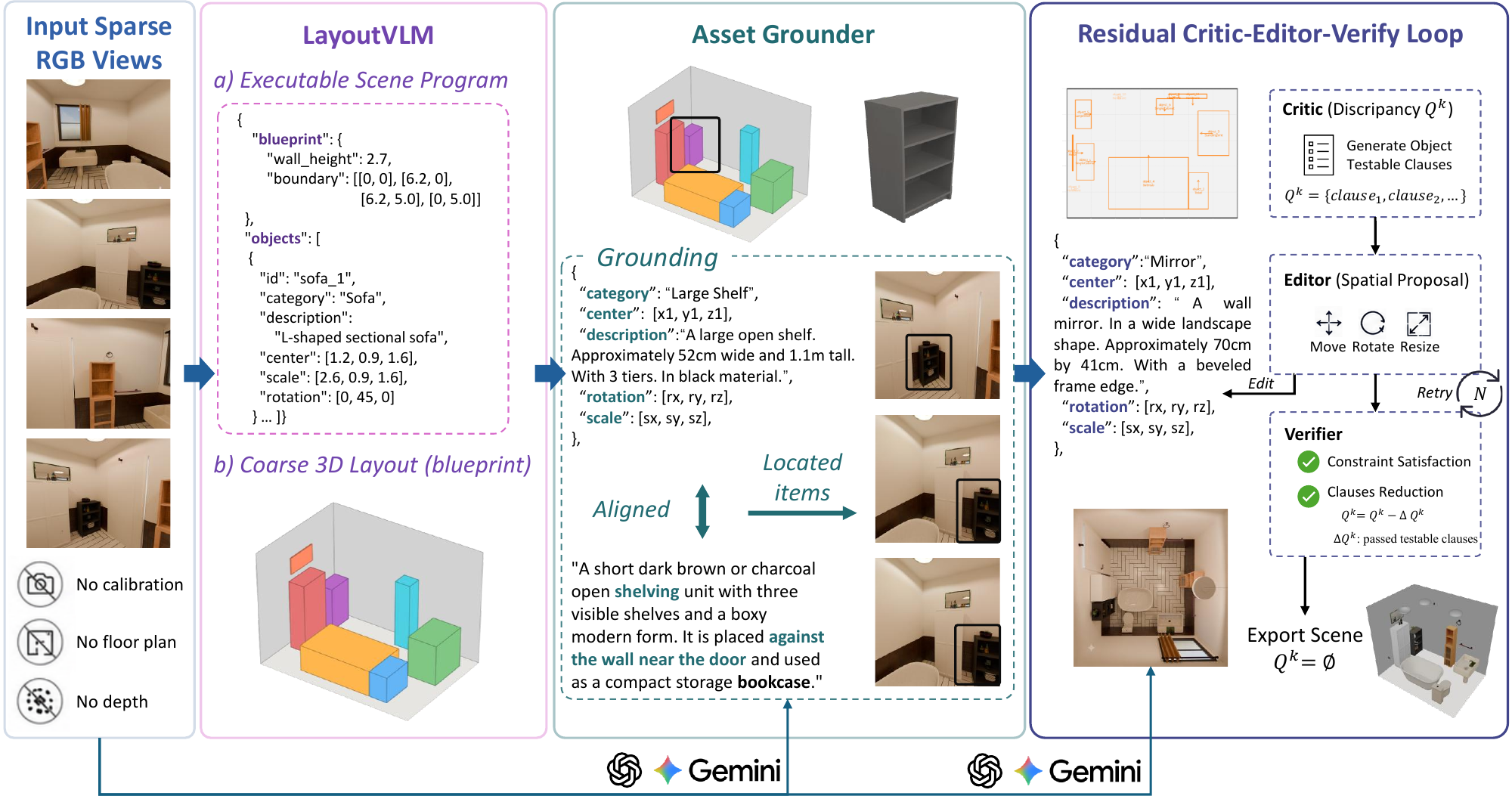}
    % \vspace{-0.2cm}
    \caption{Overview of \method. LayoutVLM predicts an executable scene program from sparse RGB observations; Asset Grounder associates observed instances with scene entities and conditions open-vocabulary asset generation, and these visual evidences together with texts will be used to generate 3D assets; Residual Critic–Editor–Verify Loop performs closed-loop refinement under deterministic geometric constraints, and finally exports the full scene when the testable clause set is empty.}
    \label{fig:pipeline}
    \vspace{-0.2cm}
\end{figure*}

\section{Related Work}
\label{sec:related-work}

\paragraph {Structured indoor scene synthesis.}
SceneFormer, ATISS, and DiffuScene learn generative priors over object configurations~\citep{wang2021sceneformer,paschalidou2021atiss,tang2024diffuscene}, while LayoutGPT, Holodeck, AnyHome, and NaLA broaden control through language, relations, or foundation-model agents~\citep{feng2023layoutgpt,yang2024holodeck,fu2024anyhome,wan2026nala}. Geometry-conditioned systems operate from proxy boxes or supplied layouts~\citep{schult2024controlroom3d,fang2026spatialgen}, and SceneWeaver and SAGE coordinate generators with semantic, visual, or physical evaluation~\citep{yang2025sceneweaver,xia2026sage}. These methods synthesize from specifications rather than recover the identity, appearance, and metric placement of instances across sparse views.

\paragraph {Observation-conditioned scene recovery.}
Architectural methods recover room envelopes from perspective or panoramic RGB and polygonal floor plans from point clouds~\citep{zou2018layoutnet,sun2019horizonnet,pintore2020atlantanet,yue2023roomformer}. Object-centric systems estimate cuboids and meshes, align CAD assets, or combine open-world perception with pose optimization~\citep{nie2020total3dunderstanding,gumeli2022roca,wu2025diorama}, but remain constrained by asset coverage and occlusion. Generative recovery reduces library dependence but still requires masks, estimated geometry, de-occlusion, or separate pose stages~\citep{meng2026scenegen,yin2026threedfixer,shi2026scenemaker}; SceneScript decodes structured commands from walkthrough video~\citep{avetisyan2024scenescript}. Sparse-view radiance fields assume calibrated cameras~\citep{niemeyer2022regnerf,yang2023freenerf}; DUSt3R and MASt3R align pointmaps and dense correspondences~\citep{wang2024dust3r,leroy2024mast3r}, while Spann3R, CUT3R, and VGGT predict cameras, depth, pointmaps, and tracks through spatial memory, recurrence, or feed-forward inference~\citep{wang2025spann3r,wang2025cut3r,wang2025vggt}. These approaches recover geometry rather than named, independently regenerable entities with editable relations. We instead predict an object-addressable scene program preserving category, appearance, and cross-view identity from sparse uncalibrated views. 

% \paragraph{Asset realization, spatial reasoning, and supervision.}
% CRM, One-2-3-45, Unique3D, and TRELLIS reconstruct meshes or structured 3D representations from generated multiview, image, or text conditions~\citep{wang2024crm,liu2023one2345,wu2024unique3d,xiang2025trellis}, but do not determine scene inventory or placement. Spatial foundation models learn metric relations from spatial-VQA, region prompts, point clouds, or image-derived 3D features~\citep{chen2024spatialvlm,cheng2024spatialrgpt,hong2023threedllm,chen2024ll3da}; related systems optimize VLM-derived layouts or plan validated edits to existing scenes~\citep{sun2025layoutvlm,noh2026editasact}. Supervision remains fragmented: SUN RGB-D provides single RGB-D frames and oriented boxes, ScanNet posed RGB-D video and surfaces, Structured3D synthetic architectural annotations, and 3D-FRONT furnished rooms~\citep{song2015sunrgbd,dai2017scannet,zheng2020structured3d,fu2021threedfront}. None aligns sparse uncalibrated RGB observations with executable room-and-object records containing generative descriptions and full placement parameters. I3SL fills this gap, making scene structure both a direct prediction target and an editable interface for asset realization. This joint target connects predicted geometry to replaceable assets and explicit relations without rerunning reconstruction.

\section{Method}
\label{sec:method}

\subsection{Problem Formulation}
\label{sec:problem-formulation}

Given sparse RGB observations $\mathcal{I}=\{I_v\}_{v=1}^{V}$ of an indoor environment, our goal is to recover a structured and editable 3D scene without externally provided camera calibration, depth, point clouds, floor plans, or object masks. Conventional object-centric pipelines decompose this task into detection, geometric lifting, cross-view association, and optimization; we instead formulate reconstruction as \emph{direct conditional generation of a 3D scene program}.

We represent a scene as $\mathcal{S}=(\mathcal{B},\mathcal{O},\mathcal{R})$. LayoutVLM first predicts an initial program $\widehat{\mathcal{S}}_0=(\mathcal{B},\widehat{\mathcal{O}}_0)$, where $\mathcal{B}$ denotes the induced blueprint used by subsequent stages; visual evidence then grounds the object set and relations. The blueprint $\mathcal{B}$ specifies polygonal room boundaries and wall parameters, while $\mathcal{O}=\{o_i\}_{i=1}^{N}$ contains individually addressable entities:
\begin{equation}
    o_i=(c_i,d_i,\mathbf{t}_i,\mathbf{s}_i,\mathbf{r}_i,u_i),
\end{equation}
where $c_i$ and $d_i$ denote category and appearance, $\mathbf{t}_i\in\mathbb{R}^{3}$ and $\mathbf{s}_i\in\mathbb{R}_{+}^{3}$ are the 3D center and extent, $\mathbf{r}_i$ is the orientation, and $u_i$ records floor or object support. The relation set $\mathcal{R}$ captures object--object and object--room constraints. This representation is predictive and executable: it can be rendered as a canonical spatial proxy, edited explicitly, and populated with independently generated assets.

Figure~\ref{fig:pipeline} summarizes the pipeline. LayoutVLM predicts an initial scene program from all views in one step; Asset Grounder matches its entities to visual evidence and generates assets; the residual loop refines the spatial arrangement.

\subsection{Direct 3D Scene Program Induction}
\label{sec:scene-prediction}

\emph{Vision-to-structure generation.}
LayoutVLM shifts the task from staged geometric estimation to direct structured generation. Given a variable number of views in a single multimodal context, it autoregressively emits a serialized scene program rather than constructing per-view detections and lifting them into 3D. It jointly predicts room structure, object identities, semantic descriptions, and metric transformations:
\begin{equation}
    \widehat{\mathcal{S}}_{0}
    = \arg\max_{\mathcal{S}_0}
      p_{\theta}\!\left(\mathcal{S}_0\mid I_1,\ldots,I_V\right).
    \label{eq:direct-induction}
\end{equation}
Here, $p_\theta$ is LayoutVLM's serialized-program distribution with parameters $\theta$.
The constrained schema separates blueprint and object records, each coupling semantic and appearance information with metric placement. It therefore expresses decisions normally made in separate coordinate systems within one compositional output space, while allowing weakly overlapping or non-overlapping views to contribute complementary evidence without explicit correspondence.
% Each object record couples its category and description with a 3D center, extent, and orientation parameters. Consequently, structural decisions that conventional pipelines make in separate coordinate systems are expressed in one compositional output space. This direct formulation also allows non-overlapping observations to contribute semantic evidence even when explicit feature correspondence is unavailable.

\emph{Observation-consistent learning.}
A sampled view subset rarely observes every entity, so supervising hidden objects would conflate reconstruction with unconstrained completion. We form $\mathcal{O}^{\mathrm{obs}}$ by retaining an object when the number of its oriented-box corners visible across the selected views reaches a fixed threshold $\kappa$; a corner is visible only if it lies inside a camera frustum and its ground-plane sight line does not intersect a blueprint wall. Camera parameters are used only to precompute targets and are unavailable at inference. For $\mathcal{S}_0^{\mathrm{obs}}=(\mathcal{B}^{\mathrm{gt}},\mathcal{O}^{\mathrm{obs}})$ and $\mathbf{y}^{\mathrm{obs}}=\operatorname{Ser}(\mathcal{S}_0^{\mathrm{obs}})$, LayoutVLM minimizes
\begin{equation}
    \mathcal{L}_{\mathrm{ind}}
    =-\sum_{t=1}^{T}
      \log p_{\theta}\!\left(
      y_t^{\mathrm{obs}}\mid y_{<t}^{\mathrm{obs}},\mathcal{I}\right).
    \label{eq:layout-loss}
\end{equation}
Here, $\mathcal{B}^{\mathrm{gt}}$ is the annotated blueprint, $\operatorname{Ser}$ deterministic serialization, and $T$ the target length; only assistant tokens are supervised, with user and system tokens masked. Targets are generated offline from random view subsets.
% where loss is applied only to assistant output tokens. This aligns direct structural induction with the evidence present in the input rather than rewarding unsupported hallucination. To accelerate training, we integrated this part into the dataset preprocessing stage as offline data augmentation.

\subsection{Asset Grounder}
\label{sec:asset-grounding}

Direct induction provides a global spatial hypothesis, but ambiguous or out-of-distribution observations may still cause omissions, duplicates, or category errors. We therefore separate two authorities: the scene program supplies the room frame and initial poses, while visual evidence determines which instances exist and how they appear.

We construct an evidence-backed multiview inventory $\mathcal{E}=\{e_j\}_{j=1}^{M}$. Joint and per-view reasoning resolve cross-view identity, while a count audit handles repeated instances. Each item records category, appearance, support type, confidence, and evidence views.

An MLLM reconciles $\mathcal{E}$ with $\widehat{\mathcal{O}}_0$ using semantic and coarse spatial cues. Each seed receives one retention, replacement, or suppression decision, whereas cloning is reserved for an unmatched inventory item. These operations remove unsupported seeds, correct categories while retaining useful poses, and restore missing repeated instances without expanding $\mathcal{E}$. The grounded set $\widehat{\mathcal{O}}_g=\{o_i^g\}_{i=1}^{M}$ satisfies:
\begin{equation}
\begin{aligned}
    \widehat{\mathcal{O}}_{g}
    &=\operatorname{Recon}\!\left(
      \widehat{\mathcal{O}}_0,\mathcal{E}\right),\\
    |\widehat{\mathcal{O}}_g|&=M,\qquad
    \pi:\mathcal{E}\rightarrow\widehat{\mathcal{O}}_g
    \ \text{is bijective},\\
    \operatorname{fam}(e_j)&=\operatorname{fam}(\pi(e_j)),
    \quad \forall e_j\in\mathcal{E}.
\end{aligned}
\label{eq:grounded-reconciliation}
\end{equation}
Here, $\operatorname{Recon}$ is the reconciliation operator, $\pi$ maps inventory items to active entities, and $\operatorname{fam}$ denotes semantic family. The bijection maps every inventory item to exactly one active entity. Reconciliation predicts no new poses; an image-guided planner localizes clones before boundary and collision checks validate all poses.

The grounded program determines \emph{where} each object belongs, while observations determine \emph{how} it appears. A promptable segmenter proposes regions and a VLM selects the best soft mask. The mask and description condition a clean single-object reference image, with text-only generation used when no mask is reliable. An image-conditioned 3D generator converts this reference into a mesh, while the program extent controls fitting during assembly.

\subsection{Residual Critic--Editor--Verify Loop}
\label{sec:agentic_refinement}
Grounding and asset insertion may leave residual position, scale, and facing errors due to uncertain placement, metric size, or local front conventions. We therefore use a bounded Critic--Editor--Verify loop to refine, rather than reconstruct, the grounded layout.
% Modern 3D generative models can faithfully recover appearance details, but the generated assets often come with ambiguous metric scale and uncertain canonical orientation. When such assets are instantiated into a metric 3D layout, these ambiguities directly cause practical integration failures, \eg wrong facing directions and mismatched sizes, even if the predicted layout geometry is correct. We therefore introduce a Residual Critic--Editor--Verify loop to iteratively reduce these geometric residuals.

Starting from $\mathcal{S}^{0}=\widehat{\mathcal{S}}_g$, we render a metric bird's-eye proxy $\mathcal{P}(\mathcal{S}^{k})$ with footprints and facing directions. An image generator produces an auxiliary top-down hypothesis $\mathcal{H}(\mathcal{I})$. Because $\mathcal{H}$ is synthesized, it supplies only coarse position and wall cues; the metric program and high-confidence relations remain authoritative.
% We render the perdicted 3D layout ${\widehat{\mathcal{S}}_{g}}$ into a metric bird's-eye proxy with object footprints and facing arrows. An image generator also produces an auxiliary top-down hypothesis from the sparse photographs by composing the grounded visual information, these results are extracted in the last step of Sec.~\ref{sec:asset-grounding}. Because this image is generated rather than measured, it supplies only coarse position and wall evidence; the relation graph and orientation plan remain authoritative for semantic facing.

At round $k$, the Critic converts visible discrepancies into object-specific clauses $\mathcal{Q}^{k}$, the Editor proposes a minimal translation, rotation, or footprint change, and the Verifier judges every clause on a fresh rendering:
\begin{equation}
\begin{aligned}
\mathcal{Q}^{k}
    &=\operatorname{Critic}\!\left(
      \mathcal{P}(\mathcal{S}^{k}),\mathcal{H}(\mathcal{I}),\mathcal{R}\right),\\
\mathcal{S}^{k+1}
    &=\operatorname{Edit}(\mathcal{S}^{k},\mathcal{Q}^{k}),\\
\mathbf{a}^{k+1}
    &=\operatorname{Verify}\!\left(
      \mathcal{P}(\mathcal{S}^{k+1}),
      \mathcal{H}(\mathcal{I}),\mathcal{Q}^{k},\mathcal{R}\right).
\end{aligned}
\label{eq:cev-update}
\end{equation}
Each entry of $\mathbf{a}^{k+1}$ is a pass/fail verdict.
Failed clauses and verifier feedback guide the next edit, while a deterministic footprint test introduces blueprint violations as mandatory constraints.
Each edit is checked against the scene schema before application.
The loop terminates when all clauses pass and all footprints lie within the blueprint, when no valid edit is applied, or when its fixed iteration budget is exhausted.

\section{The \dataset{} Dataset}
\label{sec:dataset}
\begin{figure*}[t]
    \centering
    \includegraphics[width=\textwidth]{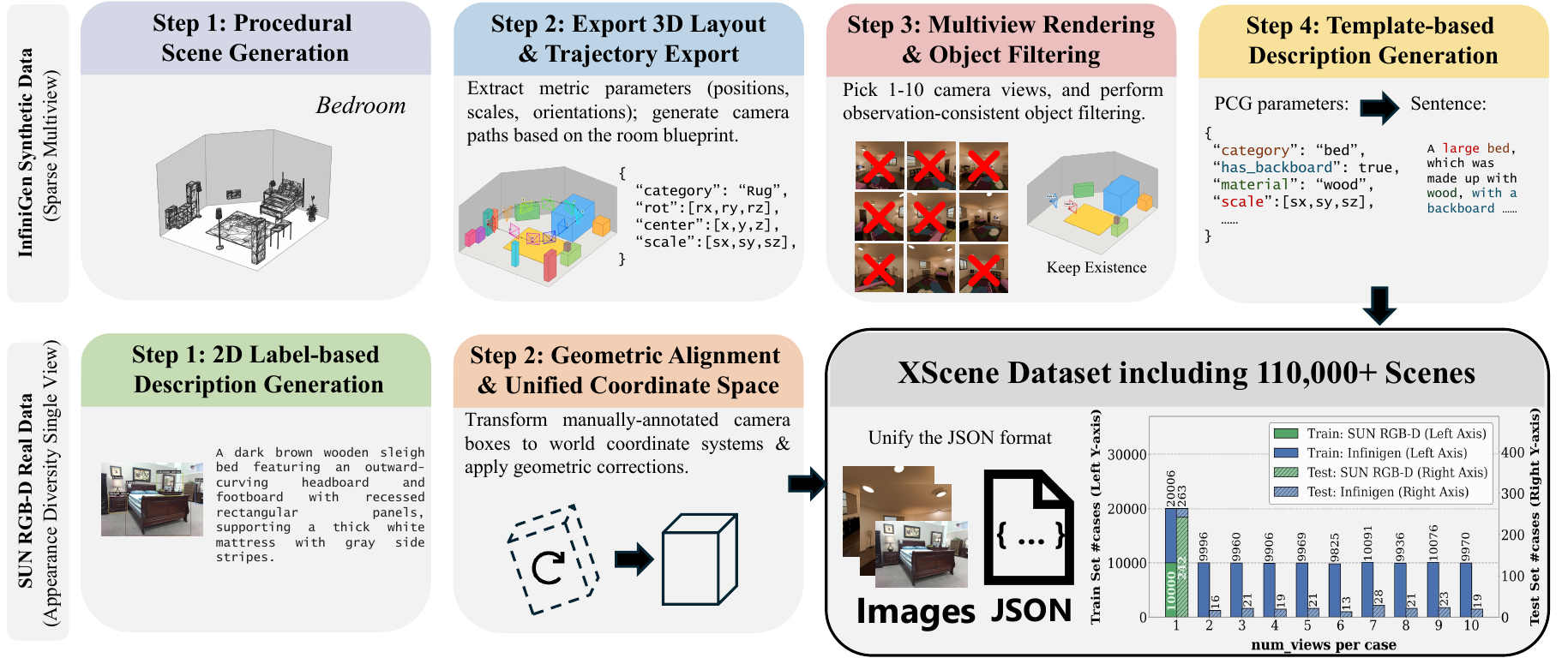}
    \caption{Construction of \dataset. Synthetic and real-image sources provide complementary appearance and structural diversity, while cross-view visibility aggregation aligns each supervision target with the evidence available in its selected observations.}
    \label{fig:dataset-overview}
\end{figure*}

% 构造用于训练LayoutVLM的\dataset{}数据集，该数据集需要为模型训练提供: 1) perpective views 2) objects 与 blueprint 的 3D layout。其中每个object需要包含longform descriptions, metric positions, extents, and orientations. 我们通过InfiniGen程序化随机生成合成数据集场景，以及对现有的3d数据集进行补充标注.
% We constructed the \dataset{} dataset for training LayoutVLM. This dataset must provide the following for model training: 1) perspective view images, and 2) 3D layouts of objects and blueprints. Each object needs to include long-form descriptions, metric positions, extents, and orientations. To obtain the required data, we procedurally generate randomized synthetic scenes using InfiniGen and perform supplementary annotations on an existing 3D dataset, SUN RGB-D.
Training LayoutVLM requires paired RGB observations from different perspectives and structured 3D scene programs. Each target must encode a room blueprint together with each object's long-form description, metric position, extent, and orientation. Since existing indoor datasets lack this complete supervision, we construct the Image--3D Scene Layout dataset (\dataset) by procedurally generating randomized multiview scenes with InfiniGen and supplementing SUN RGB-D with the missing object annotations (Fig.~\ref{fig:dataset-overview}).

% Double-column dataset overview placeholder.

% \subsection{Preliminary of InfiniGen}
% InfiniGen is a procedural indoor scene generation framework that synthesizes diverse photorealistic environments through procedural content generation (PCG). Given a room type, it first generates the room structure, including walls, doors, and windows, and then procedurally creates and arranges furniture according to functional layout priors. Since every scene is generated from explicit procedural representations, InfiniGen can directly provide accurate object-level annotations, including semantic categories, 3D positions, extents, orientations, and room layouts. These characteristics make it a suitable platform for constructing large-scale datasets with precise structural supervision. The whole generation progress is controlled by a random seed. 
% 2

\subsection{Data Construction}
\label{sec:data-construction}
% Describe the synthetic Infinigen scenes and the real SUN RGB-D subset.
% \textbf{Infinigen part.}
\paragraph{Procedural multiview data.}
Using fixed InfiniGen seeds, we generate diverse indoor scenes across room types, such as kitchens, bathrooms, dining rooms, bedrooms, and living rooms, and export their room envelopes and object categories, positions, extents, and orientations. For each room, a blueprint-aware camera trajectory renders ten RGB images, from which we randomly retain one to ten views. Frustum intersection and wall occlusion retain only objects supported by the selected observations. InfiniGen's 79 procedural object generators expose readable parameters, which category-specific templates convert into long-form appearance descriptions.
% We use fixed seeds to generate indoor scenes (\eg kitchens, bathrooms, dining rooms, bedrooms, and living rooms). Furniture is then generated and placed into these spaces. The metric positions, extents, and orientations of the objects are directly exported from these scenes. Next, for each room, a fixed camera trajectory is generated based on the shape of the blueprint layout to render 10 RGB images. These images are randomly dropped, retaining only 1 to 10 views per room. To account for occlusions, the selected perspective views are further used to filter the objects, keeping only those visible inside the room. We use Nano-Banana to enhance the reality of RGB images. 

\paragraph{Real-image data.}
Although these generators can theoretically yield unlimited assets, fixed sentence templates and parameterized property controls often produce structurally similar objects that underrepresent real-world diversity. We therefore add SUN RGB-D as single-view real-image supervision and use its 2D labels to prompt Hunyuan-turbo to describe the corresponding objects, attaching the text to their original 3D annotations. We transform the 3D boxes from camera coordinates to a world frame with non-negative $x$, $y$, and $z$ coordinates. Because these boxes are manually annotated and may contain pose errors, we geometrically correct cases such as floor-supported objects whose boxes do not align with the ground plane. Finally, we convert the processed SUN RGB-D and InfiniGen subsets to a unified scene-program schema.
% As objects in the InfiniGen indoor are generated by 79 randomized procedural object generators, and each generator is driven by readable keys and parameters, \eg $has\_backboard=false$, $backboard\_thickness=0.016$. We tailored different sentence templates for each assert generator, which can convert a PCG-style parameter into descriptive sentences. 

% \textbf{SUN RGB-D part.} In the InfiniGen subset, object descriptions are generated using fixed templates. Although the procedural generators can theoretically produce an infinite variety of assets, parameterized property controls often yield objects with similar structural characteristics, which fail to fully capture real-world diversity. To address this limitation, we incorporate and process the SUN RGB-D dataset.

% First, leveraging the 2D image labels of the original dataset, we employ the Hunyuan-turbo model to generate textual descriptions for the corresponding objects, which are subsequently integrated into the 3D annotations. Next, we transform these 3D labels from the camera coordinate system to the world coordinate system, shifting the coordinate space to non-negative intervals along the x, y, and z axes. Furthermore, because the 3D bounding boxes in SUN RGB-D are manually annotated, they frequently exhibit pose inaccuracies. \eg the bounding box of an object resting on the floor may not align perfectly with the ground plane. We apply geometric corrections to rectify these poses. Finally, the processed SUN RGB-D and InfiniGen datasets are consolidated into a unified format.

\subsection{Dataset Statistics}
\label{sec:dataset-statistics}

% \begin{figure}[t]
%     \centering
%     \begin{subfigure}[t]{0.49\linewidth}
%         \includegraphics[width=\linewidth]{figures/view_dist_merged_train.pdf}
%         \caption{Training split.}
%         \label{fig:view-dist-train}
%     \end{subfigure}
%     \hfill
%     \begin{subfigure}[t]{0.49\linewidth}
%         \includegraphics[width=\linewidth]{figures/view_dist_merged_test.pdf}
%         \caption{Test split.}
%         \label{fig:view-dist-test}
%     \end{subfigure}
%     \caption{Per-case view-count distribution of \dataset{}. SUN RGB-D contributes
%     single-view cases (green); Infinigen cases (blue) span $1$--$10$ views.}
%     \label{fig:view-dist}
% \end{figure}

% \begin{table}[t]
%     \centering
%     \caption{Per-split statistics of \dataset{}. A \emph{case} denotes a room paired
%     with its perspective views.}
%     \label{tab:\dataset{}-splits}
%     \small
%     \setlength{\tabcolsep}{4pt}
%     \begin{tabular}{l l r r c}
%     \toprule
%     Subset & Split & \#Cases & \#Images & Views / case \\
%     \midrule
%     SUN RGB-D  & train & 10{,}000  & 10{,}000  & 1 \\
%     SUN RGB-D  & test  & 242       & 242       & 1 \\
%     Infinigen  & train & 99{,}735  & 548{,}806 & 1--10 \\
%     Infinigen  & test  & 202       & 1{,}136   & 1--10 \\
%     \midrule
%     \dataset{}       & total & 110{,}179 & 560{,}184 & 1--10 \\
%     \bottomrule
%     \end{tabular}
% \end{table}

We split \dataset{} at the case level so that no room appears in both training and test. Table in \ref{fig:dataset-overview} reports 110,179 cases and 560,184 images in total; InfiniGen provides scale and dense multiview supervision ($\sim\!91\%$ of training cases), while SUN RGB-D contributes real-world imagery and human-verified 3D annotations. InfiniGen cases are approximately uniformly distributed across one to ten views by construction, whereas SUN RGB-D occupies the single-view bin, together spanning the full spectrum from sparse to dense observations.
\section{Experiments}
\label{sec:experiments}
% Double-column qualitative comparison placeholder.
\begin{figure*}[t]
    \centering
    \includegraphics[width=\linewidth]{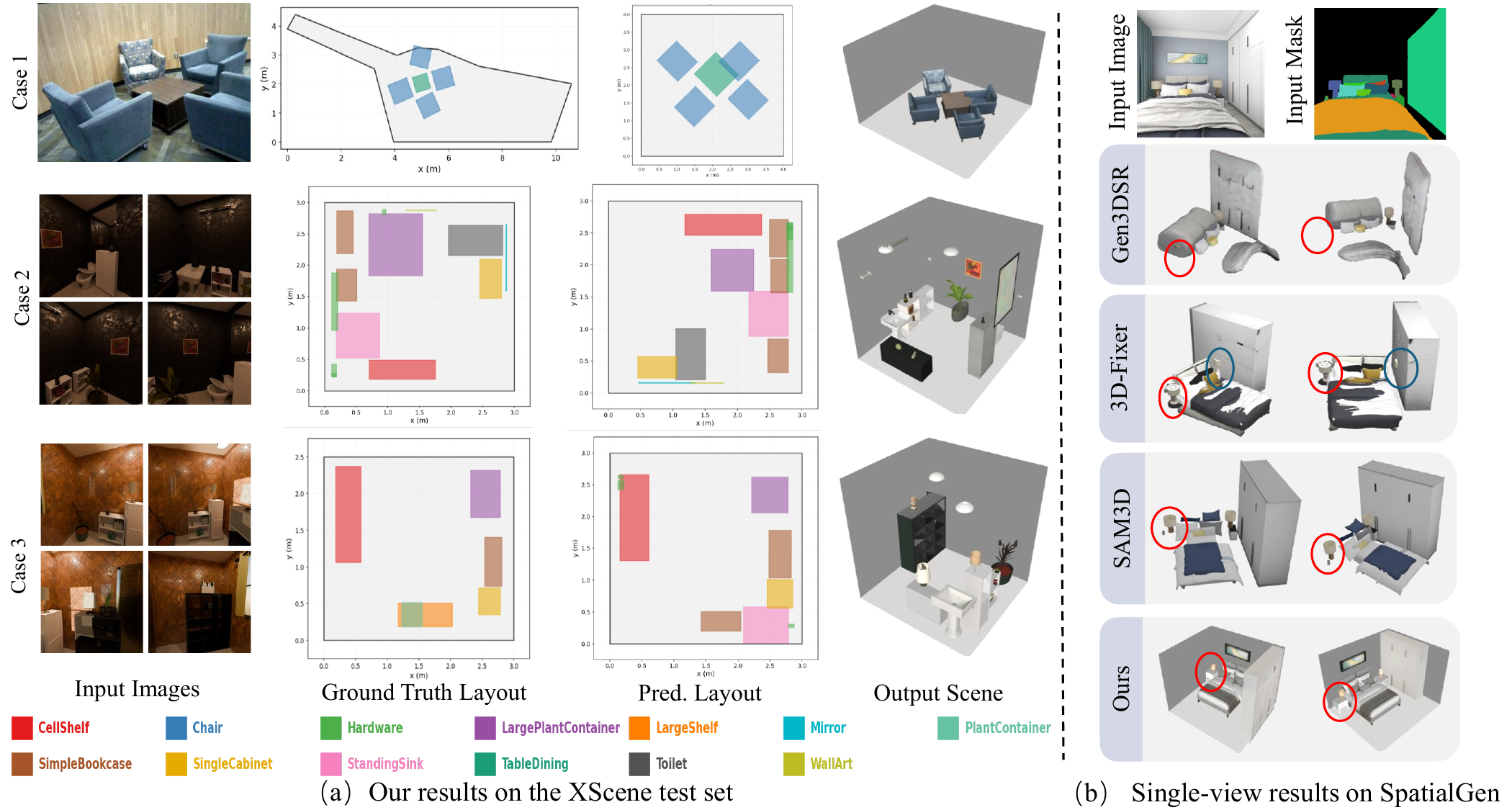}
    \caption{Qualitative results of \method{} across in-domain, real-image, and out-of-distribution scenes. }
    \label{fig:qualitative-results}
    \vspace{-1mm}
\end{figure*}

\subsection{Experimental Setup}
\label{sec:experimental-setup}
% Datasets, baselines, implementation details, and evaluation protocol.
We train Qwen3.5-4B and Qwen3.5-9B LayoutVLMs by supervised fine-tuning. The \dataset{} test split contains 202 InfiniGen cases with one to ten views and 242 single-view SUN RGB-D cases. For OOD evaluation, we subsample three to ten of each SpatialGen scene's 100 continuous views~\cite{fang2026spatialgen}; a single-view subset supports comparisons with SAM-3D-object (SAM3D)~\cite{chen2026sam3d}, Gen3DSR~\cite{ardelean_gen3dsr_2025}, and 3D-Fixer~\cite{yin2026threedfixer}.

\subsection{Evaluation Metrics}
Predictions and ground truth use independent coordinate frames, so we remove global translation and rotation before distance evaluation. 
% We anchor alignment on the largest-area furniture class appearing exactly once in both scenes, falling back to the largest shared class, and search a discrete rotation--reflection--translation grid. Coordinates are normalized by each room diagonal, making distance thresholds scale invariant.

We report four complementary metric families.
\textbf{(1) Localization F1.} We greedily match predicted and ground-truth objects of the same class by normalized center distance, counting a true positive below $\tau$. Precision, Recall, and F1 are reported at $\tau\in\{5\%,10\%\}$ of the room diagonal using macro scene averages and micro pooled counts.
\textbf{(2) Class-only Metrics.} Ignoring position, we measure class-count overlap with $\mathrm{TP}=\sum_c\min(n_c^{\mathrm{pred}},n_c^{\mathrm{gt}})$. Its gap from localization F1 separates recognizing \emph{what} belongs in a room from placing it \emph{where} it belongs.
\textbf{(3) 3D IoU.} We compute oriented-box IoU for matched pairs and report its mean over matches, its mean over all ground-truth instances with misses counted as zero, and F1@\{0.25,0.5\}. Class-aware matching requires equal classes; class-agnostic matching provides a localization upper bound. 
\textbf{(4) Orientation Error (ICP-Rot).} Symmetry-aware yaw error folds the residual angle by each footprint's rotational symmetry ($180^\circ$ for rectangular and $90^\circ$ for near-square footprints), avoiding penalties for equivalent orientations.

% \textbf{(5) View-Consistent 2D Metrics.} For the single-view subset,
% where a calibrated camera is available, we additionally project
% matched 3D boxes into the actual image plane and compute (i) per-object
% 2D silhouette IoU and F1@\{0.25, 0.5\}, and (ii) a scene-level
% \emph{Union-IoU} that rasterizes all foreground objects into a single
% mask on each side and reports the overall silhouette overlap,
% independent of per-object class matching. For qualitative and
% perceptual comparison we further render the predicted scene from the
% same camera and report LPIPS and RMSE against the input photograph,
% restricted to the ground-truth foreground region.
Unless noted otherwise, a fixed class-pattern filter excludes small decorative objects such as lights, tableware, books, wall ornaments, and hardware, which add disproportionate noise without reflecting room-scale layout errors.

\subsection{Structured Scene Prediction}
\label{sec:scene-prediction-results}

\begin{table*}
\centering
\small
\caption{Raw structured scene prediction on in-domain (Infinigen,
SUN RGB-D) and out-of-domain (SpatialGen) splits, macro-averaged over
5--6 inference runs (mean\,$\pm$\,std). Cls.\ P/R/F1 is the
location-agnostic class-count-matching precision/recall/F1
(micro-averaged); it disentangles semantic recognition from spatial
placement.}
\label{tab:main-results}
{\setlength{\tabcolsep}{3.5pt}% tighter columns (local)
\renewcommand{\arraystretch}{0.95}% slightly tighter rows (local)
\resizebox{\textwidth}{!}{%
\begin{tabular}{ll ccc ccc c c}
\toprule
\textbf{Model} & \textbf{Dataset} & \textbf{F1@5\%$\uparrow$} & \textbf{F1@10\%$\uparrow$} & \textbf{Cls.P$\uparrow$} & \textbf{Cls.R$\uparrow$} & \textbf{Cls.F1$\uparrow$} & \textbf{3D-IoU$_{\text{match}}$$\uparrow$} & \textbf{ICP-Rot($^\circ$)$\downarrow$} \\
\midrule
4B & Infinigen (in-domain) & $0.299\pm.003$ & $0.471\pm.002$ & $0.858\pm.002$ & $\mathbf{0.801}\pm.001$ & $\mathbf{0.822}\pm.001$ & $\mathbf{0.267}\pm.003$ & $28.3\pm0.8$ \\
9B & Infinigen (in-domain) & $\mathbf{0.331}\pm.003$ & $\mathbf{0.511}\pm.002$ & $\mathbf{0.889}\pm.002$ & $0.758\pm.002$ & $0.800\pm.002$ & $0.265\pm.006$ & $\mathbf{26.5}\pm1.9$ \\
\midrule
4B & SUN RGB-D (in-domain) & $0.515\pm.005$ & $0.601\pm.007$ & $\mathbf{0.684}\pm.005$ & $0.655\pm.003$ & $0.635\pm.003$ & $0.420\pm.004$ & -- \\
9B & SUN RGB-D (in-domain) & $\mathbf{0.519}\pm.007$ & $\mathbf{0.620}\pm.005$ & $0.678\pm.002$ & $\mathbf{0.697}\pm.003$ & $\mathbf{0.653}\pm.003$ & $\mathbf{0.427}\pm.002$ & -- \\
\midrule
4B & SpatialGen (OOD, raw) & $0.213\pm.007$ & $0.288\pm.009$ & $0.332\pm.008$ & $\mathbf{0.697}\pm.020$ & $0.413\pm.007$ & $0.214\pm.009$ & -- \\
9B & SpatialGen (OOD, raw) & $\mathbf{0.239}\pm.009$ & $\mathbf{0.333}\pm.014$ & $\mathbf{0.428}\pm.018$ & $0.677\pm.019$ & $\mathbf{0.477}\pm.015$ & $\mathbf{0.217}\pm.014$ & -- \\
\bottomrule
\end{tabular}}% end resizebox
}% end local group
\vspace{-1mm}
\end{table*}

Table~\ref{tab:main-results} reports raw, pre-refinement predictions on the two in-domain splits and zero-shot transfer to 48 OOD SpatialGen scenes. Results are macro-averaged over five runs and reported as mean $\pm$ population standard deviation where available.

\textbf{In-domain behavior.} SUN RGB-D produces substantially higher localization F1 and matched 3D IoU than InfiniGen, plausibly because its single-view scenes are smaller and less cluttered under normalized distance thresholds. InfiniGen instead retains stronger class-level coverage, consistent with its dominant share of the synthetic training distribution. These contrasting trends show that semantic inventory recovery, center localization, and volumetric agreement capture distinct errors. On InfiniGen, \textsc{9B} also lowers yaw error, indicating better orientation reasoning despite similar matched IoU.

% \textbf{In-domain performance.} SUN RGB-D achieves higher localization than InfiniGen: F1@5\% is $0.515$--$0.519$ versus $0.299$--$0.331$, possibly because its smaller, less cluttered rooms reduce ambiguity under a normalized radius. Conversely, InfiniGen has higher class-only recall ($0.800$--$0.822$ versus $0.635$--$0.653$), indicating stronger semantic coverage on the synthetic training distribution. SUN RGB-D produces higher matched 3D IoU ($0.420$--$0.427$ versus $0.265$--$0.267$), confirming that semantic coverage, center localization, and volumetric agreement capture distinct errors. ICP-Rot is available for InfiniGen, where \textsc{9B} reduces yaw error from $28.3^\circ$ to $26.5^\circ$.

\textbf{Model scale.} Moving from \textsc{4B} to \textsc{9B} improves F1@5\% and F1@10\% on all splits, with the clearest gains on InfiniGen and consistent gains on SpatialGen. The improvement is not uniform: on InfiniGen, \textsc{4B} retains slightly stronger class-level performance and matched 3D IoU, whereas \textsc{9B} lowers orientation error. Capacity therefore primarily improves spatial localization rather than every aspect of reconstruction.
% 从实验结果看，3D IoU是比Localization F1更加严格的指标，因为他会严格地惩罚物体在定位时，xyz三个方向上的误差，这对于室内场景中细长含薄边的物体（eg.wall art）的摆放评估是十分严苛的，导致该指标严重较低。
3D IoU is stricter than center-based Localization F1 because it penalizes errors along all three axes, especially for thin objects such as wall art.

\textbf{Out-of-domain transfer.} Both variants show a clear localization drop on SpatialGen, while class-level recall remains substantially higher than localization F1. Together with low raw class precision and predicted inventories of $12.2$ (\textsc{9B}) and $14.1$ (\textsc{4B}) objects against a $6.3$ ground-truth mean, this gap indicates that LayoutVLM still recognizes plausible room categories but over-generates unsupported or duplicated entities and places them less reliably under distribution shift. OOD error therefore arises from both inventory calibration and metric placement, motivating the grounded reconciliation and spatial refinement evaluated below.
% \textbf{Out-of-domain transfer (SpatialGen, raw).} Without any
% test-time refinement, raw LayoutVLM predictions on SpatialGen achieve
% F1@5\% of $0.244$ (\textsc{9B}) and $0.234$ (\textsc{4B}), a
% $\sim$25--30\% relative drop from the in-domain Infinigen numbers,
% as expected for a fully held-out furniture/room distribution. Notably,
% raw \emph{recall} is comparatively high ($0.369$--$0.428$, even
% exceeding the in-domain numbers) at the cost of precision
% ($0.185$--$0.215$), because the raw decoder tends to over-generate
% objects on unfamiliar room layouts (mean $12$--$14$ predicted objects
% vs.\ $6.3$ ground-truth objects per scene); this over-generation
% behavior, and how our Stage-2 refinement corrects for it, is examined
% in Sec.~\ref{sec:ablations}.

\subsection{Ablation Studies}
\label{sec:ablations}

% \begin{table}[t]
% \centering
% \small
% \caption{Stage-wise ablation on the SpatialGen out-of-domain multi-view
% split. Stage 1 = AssetGrounder; Stage 2 = + agentic Critic--Editor--Verify loop.}
% \label{tab:spatialgen-ablation}
% {\setlength{\tabcolsep}{4pt}% tighten columns a bit (local)
% \resizebox{\columnwidth}{!}{%
% \begin{tabular}{ll ccc cc}
% \toprule
% \textbf{Model} & \textbf{Stage} & \textbf{F1@5\%$\uparrow$} & \textbf{F1@10\%$\uparrow$} & \textbf{Cls. F1$\uparrow$} & \textbf{Cls. Pre.$\uparrow$} & \textbf{Cls. Recall$\uparrow$} \\
% \midrule
% \multirow{3}{*}{9B} & raw    & 0.244 & 0.328 & 0.484 & 0.446 & \textbf{0.701}\\
%                      & stage1 & 0.311 & 0.401 & \multirow{2}{*}{\textbf{0.571}} & \multirow{2}{*}{\textbf{0.576}} & \multirow{2}{*}{0.650}\\
%                      & stage2 & \textbf{0.327} & \textbf{0.410} \\
% \midrule
% \multirow{3}{*}{4B} & raw    & 0.234 & 0.311 & 0.429 & 0.359 & \textbf{0.745}\\
%                      & stage1 & 0.308 & 0.398 & \multirow{2}{*}{\textbf{0.566}} & \multirow{2}{*}{\textbf{0.554}} & \multirow{2}{*}{0.673}\\
%                      & stage2 & \textbf{0.347} & \textbf{0.432}  \\
% \bottomrule
% \end{tabular}}% end resizebox
% }% end local tabcolsep group
% \end{table}

\begin{table}[t]
\centering
\small
\caption{Stage-wise ablation on the SpatialGen out-of-domain multi-view
split. Stage 1 = AssetGrounder; Stage 2 = + agentic Critic--Editor--Verify loop.}
\label{tab:spatialgen-ablation}
{\setlength{\tabcolsep}{4pt}% tighten columns a bit (local)
\resizebox{\columnwidth}{!}{%
\begin{tabular}{ll ccc cc}
\toprule
\textbf{Model} & \textbf{Stage} & \textbf{F1@5\%$\uparrow$} & \textbf{F1@10\%$\uparrow$} & \textbf{Cls. F1$\uparrow$} & \textbf{Cls. Pre.$\uparrow$} & \textbf{Cls. Recall$\uparrow$} \\
\midrule
\multirow{3}{*}{9B} & raw    & $0.244$ & $0.328$ & $0.484$ & $0.446$ & $\mathbf{0.701}$ \\
                    & stage1 & $0.311$ & $0.401$ & \multirow{2}{*}{$\mathbf{0.571}$} & \multirow{2}{*}{$\mathbf{0.576}$} & \multirow{2}{*}{$0.650$} \\
                    & stage2 & $\mathbf{0.327}$ & $\mathbf{0.410}$ \\
\midrule
\multirow{3}{*}{4B} & raw    & $0.234$ & $0.311$ & $0.429$ & $0.359$ & $\mathbf{0.745}$ \\
                    & stage1 & $0.308$ & $0.398$ & \multirow{2}{*}{$\mathbf{0.566}$} & \multirow{2}{*}{$\mathbf{0.554}$} & \multirow{2}{*}{$0.673$} \\
                    & stage2 & $\mathbf{0.347}$ & $\mathbf{0.432}$  \\
\bottomrule
\end{tabular}}% end resizebox
}% end local tabcolsep group
\end{table}

Table~\ref{tab:spatialgen-ablation} evaluates refinement on 48 OOD SpatialGen scenes: Stage 1 adds Asset Grounder reconciliation and deterministic repair, and Stage 2 adds the Critic--Editor--Verify loop.
% We isolate the contribution of the Stage-2 refinement pipeline
% (AssetGrounder matching is applied identically in all three settings;
% \emph{Stage 1} additionally runs deterministic geometric gating ---
% blueprint-containment and axis-snapping --- and \emph{Stage 2} further
% adds the full agentic BEV-Editor Critic--Editor--Verify loop) on the
% 48-scene SpatialGen multi-view out-of-domain split, since this is the
% setting where raw predictions are weakest and refinement headroom is
% largest.

\textbf{Stage 1 (Asset Grounder) provides the largest gain for both models.} Reconciliation, blueprint containment, and axis snapping raise F1@5\% from $0.244\to0.311$ for \textsc{9B} and $0.234\to0.308$ for \textsc{4B} ($+27.5\%$ and $+31.6\%$) without the agentic BEV-Editor; F1@10\% rises from $0.328\to0.401$ and $0.311\to0.398$. Predicted counts fall from $12.2\to6.5$ and $14.1\to6.7$, approaching the $6.3$ ground-truth mean, with roughly $5$--$7$ fewer false positives per scene. The accompanying precision gains show that unsupported, duplicated, or out-of-room entities dominate raw errors. Although pruning can reduce class recall, it improves the scene-level metrics overall.
% \textbf{Stage 1 (structured gating) recovers most of the gain, for
% both models.} Adding deterministic blueprint-containment and
% axis-snapping alone lifts F1@5\% from $0.244\to0.311$ for \textsc{9B}
% ($+27.5\%$ relative) and from $0.234\to0.347$ for \textsc{4B}
% ($+48.4\%$ relative) --- entirely without invoking the VLM-based
% editor. The mechanism is visible in the object-count statistics: mean
% predicted objects per scene drops from $12.2\to6.5$ (\textsc{9B}) and
% $14.1\to6.7$ (\textsc{4B}), essentially matching the $6.3$
% ground-truth mean, while mean false positives drop by a corresponding
% $\sim$5--7 objects/scene. This confirms our Introduction's hypothesis
% that a large share of the raw VLM's error is not fine-grained
% placement error but grossly \emph{invalid} object hypotheses
% (duplicated or out-of-room detections) that a cheap, deterministic
% geometric filter can already remove.

\textbf{Stage 2 (agentic loop) yields an additional, model-size-dependent gain.} F1@5\% improves by $+0.016$ for \textsc{9B} ($0.311\to0.327$) and $+0.039$ for \textsc{4B} ($0.308\to0.347$), while F1@10\% rises to $0.410$ and $0.432$. Class precision is unchanged because the loop edits spatial parameters rather than inventory. The full pipeline improves F1@5\% over raw predictions by $34.0\%$ and $48.3\%$: Stage 1 contributes most, while closed-loop editing consistently adds value.

%我们通过两个stage的agent优化，会删除，大量不存在于空间的物体，这的确带来了Class wise recall 层面的下降，但是对于整体的场景而言，大部分的参考指标都得到了提升。

\subsection{Single View Analysis}
\label{sec:qualitative-analysis}
% Scene visualizations, editability, and representative failure cases.
% 在这个部分，我们提供与现有方法的一些single view室内重建的方法进行对比。除了我们的模型，尽管当前并没有一个单独模型能够实现从多图，甚至是单图的全场景重建，为了体现能够与现有的单图重建模型比较3d layout的提取能力，因此，我们统一采用了Gen3DSR的mask提取方法。
We compare with reconstruction methods on the single-view SpatialGen subset. These object-centric baselines reconstruct detected instances rather than predict a complete, editable scene program from an image alone. For a fair layout comparison, all use Gen3DSR's 2D front end: CropFormer~\cite{qi2022cropformer} produces panoptic masks, OneFormer~\cite{Jain_2023_oneformer} separates foreground and background, and OVSAM~\cite{yuan2024ovsam,yuan2024mamba} assigns labels.

\begin{table}
    \centering
    \small
    \setlength{\tabcolsep}{3.5pt} % 缩小列间距以适配单栏
    \renewcommand{\arraystretch}{1.15}
    \caption{Comparison on 3D layout extraction across single-view methods. }
    \resizebox{\columnwidth}{!}{%
    \begin{tabular}{lccccccc}
    \toprule
    \textbf{Model} & \textbf{F1@5\%} & \textbf{P@5\%} & \textbf{R@5\%} & \textbf{Cls. F1} & \textbf{Cls. P} & \textbf{Cls. R} & \textbf{3D IoU} \\
    \midrule
    SAM3D     & $0.207$ & $\mathbf{0.335}$ & $0.171$ & \multirow{3}{*}{$0.396$} & \multirow{3}{*}{$\mathbf{0.618}$} & \multirow{3}{*}{$0.292$} & $0.1102$ \\
    Gen3DSR   & $0.200$ & $0.326$ & $0.165$ & & & & $0.1340$ \\
    3D-Fixer  & $0.192$ & $0.333$ & $0.150$ & & & & $0.0456$ \\
    \hline
    Ours (9B) & $\mathbf{0.355}$ & $0.316$ & $0.463$ & $\mathbf{0.457}$ & $0.414$ & $0.578$ & $0.179$ \\
    Ours (4B) & $0.287$ & $0.209$ & $\mathbf{0.564}$ & $0.381$ & $0.282$ & $\mathbf{0.705}$ & $\mathbf{0.181}$ \\
    \bottomrule
    \end{tabular}}
    \label{tab:single_view_performance_comparison}
    \vspace{-2mm}
\end{table}

% 在 Tab.~\ref{tab:single_view_performance_comparison}，我们提供了检测模型作为前序模型对于3D layout提取的影响。首先，SAM3D，Gen3DSR和3D-Fixer依托于分割模型，实现从2D到3D场景的重建，其对3D layout的理解，从技术上，依赖于2D提取阶段的模型，因此，三种方法在只考虑Class-Level的方法表现相同。然而，2D模型对于图片的非稳定分割会直接导致3D Layout错误，\eg 一个表面颜色、纹理丰富的物体，这些分割模型往往会依据色块把本属于同一个物体的实例，识别为不同的分块，这些分块又往往会被赋予不同的语义标签，一旦这些这些语义标签出现错误，3D layout的提取效果将大幅下降，进而导致整个场景在实际应用，例如，后期编辑上出现难以应用的情况，这也是为什么SAM3D等模型在 Localization F1等参数指标上全面落后于我们的模型的原因，同时，这些算法在依赖于geomerty信息的提取中，\eg。除此之外，基于2D分割的3D结果，其本质还是对于2D分割结果的3D重建，缺乏对空间的真实理解能力，经常会出现对隐藏结构的难以预测，虽然，像SAM-3D-object这些模型在训练阶段融入了DPO等后训练算法，通过对齐偏好使得模型具有了预测物体遮挡结构的能力，但这些方法依然无法完成非同类遮挡物体的预测，我们在supportive material中使用更加详细的case分析了这些内容。相反，我们的模型直接从室内场景的layout布局开始学习，能够让模型建立更加准确的空间关系理解。
Under this shared front end, the baselines inherit identical masks, labels, and class scores. Our \textsc{9B}/\textsc{4B} variants achieve F1@5\% of $0.355/0.287$ versus $0.192$--$0.207$, localization recall of $0.463/0.564$ versus $0.150$--$0.171$, and 3D IoU of $0.179$--$0.181$ versus $0.046$--$0.134$. The gain comes with lower precision, especially for \textsc{4B}; \textsc{9B} has the highest class F1 ($0.457$), whereas \textsc{4B} favors class recall ($0.705$) over precision ($0.282$).

\textbf{Why detection-based baselines fail.} The shared front end exposes a compounding error chain. Rich textures can fragment one object into multiple masks with inconsistent labels; these errors propagate to the reconstructed inventory, while camera and monocular-depth errors accumulate during 3D lifting. The resulting scene remains constrained by its 2D detections and often misses hidden structure. Although preference-aligned models such as SAM3D can complete an occluded part of a detected object, they cannot recover a fully occluded object of another category or infer its scene-level relations. LayoutVLM instead predicts identities and spatial relations jointly as a room-level program, explaining its higher recall and geometric agreement.

\textbf{Visualizations. }Figure~\ref{fig:qualitative-results}(a) shows three \dataset{} test cases with ceilings and occluding walls removed. Predictions generally match the scenes; Case 2 (row 2) shows reflection, rotation, or diagonal-symmetry ambiguity because camera relationships are not modeled, while Case 3 shows category errors such as confusing a bookcase with a large shelf or predicting a standing sink.

Figure~\ref{fig:qualitative-results}(b) compares single-view results. Even post-trained SAM3D misses the nightstand beneath the left lamp because it lacks visible pixels, whereas \method{} recovers it. More generally, segmentation can yield plausible regions yet assign incorrect semantics, causing asset-based pipelines to omit structural entities, misplace assets, and lose editability. Direct scene-program prediction instead decouples structure from asset realization, allowing each stage to improve independently and supporting human edits.

\section{Conclusion}
\label{sec:conclusion}

% Summarize the structured scene synthesis formulation, the dataset, and
% the grounding/refinement framework. State limitations without introducing
% new experiments or claims.
We introduced \method, a framework for editable 3D indoor scene synthesis from sparse, uncalibrated RGB views without external geometric priors. It directly generates executable scene programs using LayoutVLM for structural induction, Asset Grounder for perception-grounded asset generation, and an agentic Critic--Editor--Verify loop for spatial refinement. We also constructed \dataset{}, over 110,000 scenes with observation-consistent supervision. Experiments show that \method{} bridges structured prediction and open-world 3D realization, outperforming single-view baselines while preserving object-level editability across diverse environments.

\clearpage
\bibliography{aaai2027}

% AAAI-27 requires a reproducibility checklist. Keep it after references;
% complete the responses before submission.
% \newpage
% \input{ReproducibilityChecklist.tex}
\end{document}